\documentclass[]{article}
\usepackage{lmodern}
\usepackage{amssymb,amsmath}
\usepackage{ifxetex,ifluatex}
\usepackage{fixltx2e} 
\ifnum 0\ifxetex 1\fi\ifluatex 1\fi=0 
  \usepackage[T1]{fontenc}
  \usepackage[utf8]{inputenc}
\else 
  \ifxetex
    \usepackage{mathspec}
  \else
    \usepackage{fontspec}
  \fi
  \defaultfontfeatures{Ligatures=TeX,Scale=MatchLowercase}
\fi
\IfFileExists{upquote.sty}{\usepackage{upquote}}{}
\IfFileExists{microtype.sty}{%
\usepackage{microtype}
\UseMicrotypeSet[protrusion]{basicmath} 
}{}
\usepackage[margin=1in]{geometry}
\usepackage{hyperref}
\hypersetup{unicode=true,
            pdftitle={Recommender system for learning SQL using hints},
            pdfauthor={Dejan Lavbič, Tadej Matek and Aljaž Zrnec},
            pdfborder={0 0 0},
            breaklinks=true}
\usepackage{natbib}
\usepackage{longtable,booktabs}
\usepackage{graphicx,grffile}
\makeatletter
\def\maxwidth{\ifdim\Gin@nat@width>\linewidth\linewidth\else\Gin@nat@width\fi}
\def\maxheight{\ifdim\Gin@nat@height>\textheight\textheight\else\Gin@nat@height\fi}
\makeatother
\setkeys{Gin}{width=\maxwidth,height=\maxheight,keepaspectratio}
\IfFileExists{parskip.sty}{%
\usepackage{parskip}
}{
\setlength{\parindent}{0pt}
\setlength{\parskip}{6pt plus 2pt minus 1pt}
}
\providecommand{\tightlist}{%
  \setlength{\itemsep}{0pt}\setlength{\parskip}{0pt}}
\ifx\paragraph\undefined\else
\let\oldparagraph\paragraph
\renewcommand{\paragraph}[1]{\oldparagraph{#1}\mbox{}}
\fi
\ifx\subparagraph\undefined\else
\let\oldsubparagraph\subparagraph
\renewcommand{\subparagraph}[1]{\oldsubparagraph{#1}\mbox{}}
\fi

\let\rmarkdownfootnote\footnote%
\def\footnote{\protect\rmarkdownfootnote}

\usepackage{titling}

  \title{Recommender system for learning SQL using hints}
    \pretitle{\vspace{\droptitle}\centering\huge}
  \posttitle{\par}
    \author{Dejan Lavbič, Tadej Matek and Aljaž Zrnec}
    \preauthor{\centering\large\emph}
  \postauthor{\par}
    \date{}
    \predate{}\postdate{}

\usepackage{booktabs}
\usepackage{longtable}
\usepackage{array}
\usepackage{multirow}
\usepackage[table]{xcolor}
\usepackage{wrapfig}
\usepackage{float}
\usepackage{colortbl}
\usepackage{pdflscape}
\usepackage{tabu}
\usepackage{threeparttable}
\usepackage{threeparttablex}
\usepackage[normalem]{ulem}
\usepackage{makecell}

\usepackage{amsthm}

\theoremstyle{definition}

\theoremstyle{definition}

\theoremstyle{definition}

\theoremstyle{remark}

\begin{document}
\maketitle

\begin{quote}
\textbf{Dejan Lavbič}, Tadej Matek and Aljaž Zrnec. 2017.
\href{http://dx.doi.org/10.1080/10494820.2016.1244084}{\textbf{``Recommender
system for learning SQL using hints''}},
\href{https://www.tandfonline.com/toc/nile20/current}{Interactive
learning environments \textbf{(NILE)}}, 25(8), pp.~1048 -- 1064.
\end{quote}

\section*{Abstract}\label{abstract}
\addcontentsline{toc}{section}{Abstract}

Today's software industry requires individuals who are proficient in as
many programming languages as possible. Structured query language (SQL),
as an adopted standard, is no exception, as it is the most widely used
query language to retrieve and manipulate data. However, the process of
learning SQL turns out to be challenging. The need for a computer-aided
solution to help users learn SQL and improve their proficiency is vital.
In this study, we present a new approach to help users conceptualize
basic building blocks of the language faster and more efficiently. The
adaptive design of the proposed approach aids users in learning SQL by
supporting their own path to the solution and employing successful
previous attempts, while not enforcing the ideal solution provided by
the instructor. Furthermore, we perform an empirical evaluation with
\(93\) participants and demonstrate that the employment of hints is
successful, being especially beneficial for users with lower prior
knowledge.

\section*{Keywords}\label{keywords}
\addcontentsline{toc}{section}{Keywords}

Intelligent Tutoring Systems, improving classroom teaching, interactive
learning environments, programming and programming languages,
recommender system, SQL learning

\section{Introduction}\label{introduction}

Structured query language (SQL) is over three decades old and
well-adopted standard, according to ANSI. Most of today's computing
systems depend on efficient data manipulation and retrieval using SQL.
The industry is in dire need of software developers and database
administrators, proficient in SQL. Most of the SQL language is taught at
the undergraduate level of computer science schools and other
technical-oriented institutions. However, a typical student, after
completing an introductory course of databases, will not possess the
required SQL skills \citep{prior_backwash_2004}. The primary reason for
this lies in the complexity of the language itself and in the nature of
SQL teaching.

Most typically, the students misinterpret certain concepts of the
language, such as data aggregation, joins, and filtering using
predicates. Another common problem is that the students simply forget
database schemas, table and/or attribute names while constructing a
query \citep{mitrovic_modeling_2010}. \citep{prior_backwash_2004}
recognize in their study that students have difficulties in visualizing
the result of their written query, while being graded, if no option for
executing query against the database exists. As a consequence, several
SQL teaching systems now include a visualization of the database schema,
along with the tables and attributes required in the final solution, to
help students alleviate the burden of having to remember names.
Furthermore, most of the systems also include the option for testing a
query and allow students to receive feedback on their solution. Such
approaches are merely a convenience for the student, but not a solution
to help students conceptualize fundamental parts of the SQL language
itself.

Exercises for testing student's knowledge are usually oriented in a way
to check the basic concepts of SQL language, with students receiving
lectures beforehand. But once students have to apply the taught concepts
on their own, the complexity is overwhelming, unless students have had
extensive practice. With the emergence of Intelligent Tutoring Systems
(ITS), most computer-aided education systems allow the use of hints to
aid the educational process. Some of these systems were also developed
for the SQL domain. However, most of such systems encode their knowledge
manually, using expert solutions for defining rules and constraints.
Therefore, we propose a new system, which is able to offer hints for
different steps of the SQL exercise-solving process and requires minimal
intervention from experts, making overall process of defining knowledge
base for solving SQL problems much quicker and easier, as most of the
hints are generated automatically, using past exercise solutions. In
addition, our system is able to adapt to the current state the student
is in. We also perform evaluation in an actual educational environment
to determine the efficiency of the proposed system.

The rest of this paper is organized as follows. The next section
provides an overview of ITS and approaches for learning SQL. In Section
\ref{recommender-system}, we provide an extensive description of the
proposed system with system's architecture and the process of hint
generation. In Section \ref{evaluation}, we perform an empirical
evaluation of the proposed system on a group of \(93\) participants with
diverse prior knowledge of SQL. We conclude and provide future
directions for improving our research in Section
\ref{conclusions-and-future-work}.

\section{Related work}\label{related-work}

\subsection{Review of related
approaches}\label{review-of-related-approaches}

Weak learning in any discipline can be successfully addressed by
interaction between students and tutors, skilled or even not so skilled
\citep{bloom_2_1984, chen_exploring_2011, rothman_school-based_2011}.
Empirical studies showed that the interactive dialogue that occurs
between tutor and the student, and pedagogical strategies human tutor
employs are an essential component of learning
\citep{chi_active-constructive-interactive:_2009, chi_empirically_2011, ezen-can_-context_2013, jeong_knowledge_2007, lehman_how_2012}.
Human tutoring may improve student's learning performance by up to two
standard deviations \citep{bloom_2_1984, evens_one_2006}. This
encouraged researchers to investigate ways of how computer-based
teaching environments could more closely imitate human tutors and that
is what contributed to the emergence of ITS
\citep{arnau_emulating_2014, barnes_pilot_2008, eugenio_aggregation_2005, fossati_data_2015, mitrovic_learning_1998, mitrovic_db-suite:_2004, person_evaluating_2001, vanlehn_andes_2005},
computer systems intended to interact with students and help them learn.
Several of these systems have proved to be effective, although not yet
as human tutors
\citep{bloom_2_1984, eugenio_be_2008, evens_one_2006, mitrovic_db-suite:_2004, person_evaluating_2001, vanlehn_relative_2011, vanlehn_when_2007}.

ITS differ from classical computer-aided learning. They adapt to user's
individual needs in a way that they recommend educational activities and
deliver individual feedback (positive or negative) depending on the
student's profile, which includes the student's knowledge or activities
within the course they are taking \citep{anderson_cognitive_1995}.
Through the history of computer-aided learning, many ITS evolved. The
most acclaimed ones are cognitive tutors
\citep{anderson_cognitive_1995, kunmar_model-based_2002}. Later
approaches are based on either constraint-based modelling (CBM)
\citep{melia_constraint-based_2009, mitrovic_effect_2013}, a philosophy
which helps students to learn from their errors, or they construct
student models using machine learning techniques
\citep{smith-atakan_ml_2003, stein_machines_2013} to automate the rule
generation in the construction of ITS.

Cognitive tutors rely on ACT* theory that focuses on memory processes
and cognitive modelling \citep{anderson_architecture_1983}. The emphasis
of cognitive tutors is on solving procedural problems. They model the
problem domain as a set of production rules, which map out all valid
directions for solving the problem. Cognitive tutors are based on the
method of model tracing \citep{anderson_cognitive_1995}, which means
that the system monitors the student during problem-solving and when
they deviate from the right path, the system automatically offers a
hint. Because of the many possible paths for solving particular problem,
students are often forced to return to the correct path. Cognitive
tutors are hard to build because they require sophisticated AI
programming skills, they are too restrictive and they may not suit all
problem domains. For example, in \citep{koedinger_intelligent_1997},
authors estimate that the typical time to author a system is around 10
hours per production. To define production rules for a domain as complex
as SQL, it might take several years to build a useful cognitive model.
Such effort represents a serious obstacle in building tutors for complex
domains, although the need is arguably the greatest out there. For many
years, cognitive tutors have been the dominant solution in the field of
developing ITSs. Despite many new methods developed recently, cognitive
tutors remain an important building block in a computer-aided learning.

The second approach uses CBM, a design philosophy for helping students
to learn from their errors
\citep{mitrovic_modeling_2010, mitrovic_effect_2013, mitrovic_intelligent_2007, mitrovic_constraint-based_2006}.
The basic idea of the approach is that the individual problem domain is
encoded as a collection of constraints, which represent knowledge
elements of the target domain. Constraint-based tutoring system compares
the student's solution to the constraints. If all constraints are
satisfied, then the system treats the solution as correct. Violations of
constraints indicate that the student might be lacking or
misunderstanding the principles or concepts being tough. If students
violate specific constraint, the simplest implementation of CBM system
provides them a feedback associated with that constraint. When
considering the advantages of CBM, we have to emphasize that
implementation of such ITS does not require any study of student's
errors. Therefore, there is no need to explicitly encode that kind of
errors in the form of rules or sets of misconceptions. A set of
important errors is defined implicitly by specifying the constraints. An
important representative of CBM tutoring systems, also in domain of SQL,
is SQL-Tutor \citep{mitrovic_learning_1998, mitrovic_effect_2013}.
Similar to cognitive tutors, constraint-based tutors are not easy to
build and the feedback they provide could be misleading.

Advances in the field of artificial intelligence contributed to the
emergence of machine learning and data mining. They are very promising
techniques used in today's cognitive and CBM tutoring systems, where
definition of production rules is very time and cost consuming operation
\citep{aleven_new_2009, barnes_pilot_2008, fossati_data_2015, stamper_enhancing_2011}.
To achieve certain level of usefulness of cognitive tutors and CBM
systems, it is necessary to invest a significant amount of time in
defining rules and constraints. Artificial intelligence methods can be
used, to some extent, to automatically generate collections of rules,
assuming we have enough data for specific domain to learn from. The main
goal of this approach is to enable domain experts who are not skilled
programmers, to build learning models for ITS systems. Domain experts
define rules describing particular problem domain using programming by
demonstration, while the system independently builds appropriate
programme constructs.

Student data logged during learning have also gained importance, because
sufficiently a large amount of quality data enables extraction of useful
information, which can be used to build learning system. Using this kind
of approach, former student solutions are processed to build knowledge
base for specific domain. For example, in
\citep{fossati_i_2009, fossati_supporting_2009, fossati_data_2015, fossati_learning_2008},
authors introduce iList, intelligent tutoring system that uses former
student data to build knowledge model in the domain of linked lists in
computer science. Also, the authors in \citep{barnes_pilot_2008}
introduce Hint factory, intelligent system that uses student data for
building Markov Decision Process (MDP) that represents all student
approaches to solving logical problems, and also uses the MDP to
directly generate hints.

\subsection{Problem and proposed
solution}\label{problem-and-proposed-solution}

Both cognitive tutors and CBM systems use static approaches for building
problem domain, which means that in practice these systems can be built
only by high-qualified experts who thoroughly understand the domain and
possess adequate programming knowledge and skills
\citep{razzaq_assistment_2009, stamper_enhancing_2011, stein_machines_2013}.
Using artificial intelligence methods, such as data mining and machine
learning, knowledge base can be built dynamically. Dynamic building of
knowledge base eliminates the need for domain experts and enables
teachers to be involved in building intelligent systems in spite of
their lack of understanding how these systems work. If we use the domain
model based on historical data, we get the opportunity to map student's
solution to valid historical data of their colleagues, enabling the
system to support all or at least several most common paths of solving
the specific problem.

The basis of our system for learning SQL is historical data -- past
student attempts at solving SQL-related exercises. We make use of the
information within the data itself to remove the need for an expert to
encode the solutions by hand. One of the benefits of our system is that
it builds its knowledge base automatically with no intervention of a
human at all. Students are given hints based on the correct solutions of
their colleagues from previous years. In addition, the system is able to
adapt to the student's path of solving the problem and offers a hint
precisely for the current student's solution, rather than produce a
general, undirected hint. A vital component of the system is artificial
intelligence, which performs state exploration until the best state is
found and which constructs the knowledge base using past attempts. Our
approach is similar to the approach used in iList
{[}fossati\_data\_2015{]} and Hint factory \citep{barnes_pilot_2008},
two solutions that generate knowledge base with several MDPs initially
built from past student's solutions, which are later used to provide
students with hints. In contrast to iList and Hint factory, we focus on
a very complex SQL domain and define hints as partial solutions of the
problem, which act as a hint for the next step a student should take.

\section{Recommender system}\label{recommender-system}

\subsection{Description of proposed
system}\label{description-of-proposed-system}

\begin{figure}

{\centering \includegraphics[width=0.8\linewidth]{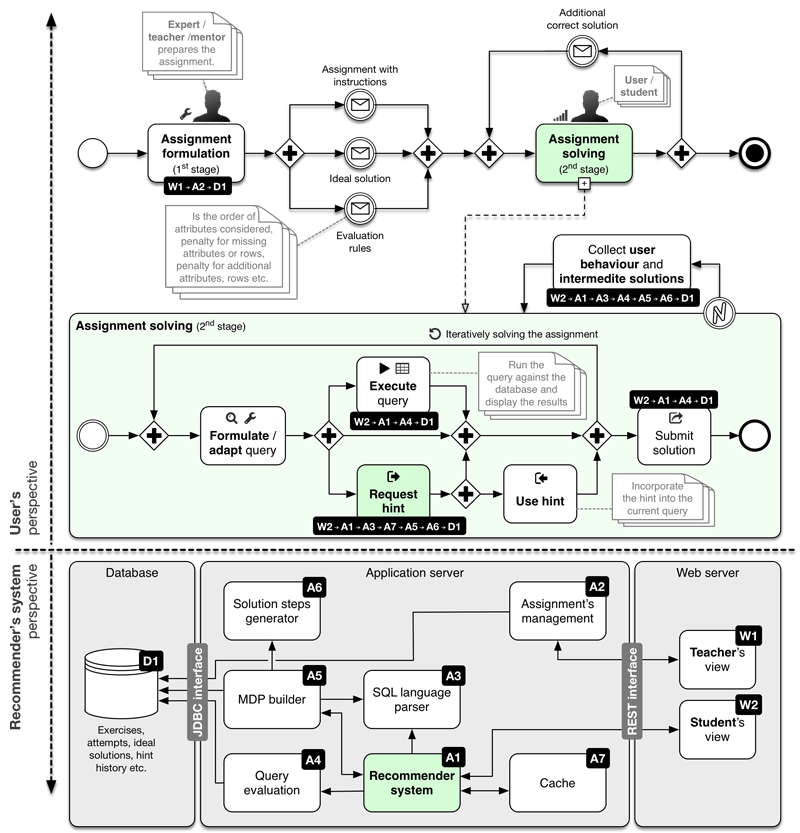}

}

\caption{Architecture of proposed recommender system from user and system perspective}\label{fig:architecture}
\end{figure}

We propose a system with intention of helping students solve SQL-related
exercises. The system is an extension of a component used in an existing
curriculum (Introduction to databases) at University of Ljubljana,
Faculty of Computer and Information Science, where students also learn
how to formulate correct and efficient SQL queries. The existing
component merely automatically evaluates the correctness of a student's
solution by comparing the result set of the student's query against the
result set of an ideal solution (instructor's solution). Points are then
deducted if certain rows or columns are missing from the result matrix
or if the order of rows is incorrect. Our system (presented in Figure
\ref{fig:architecture}) involves the application of hints as a part of
student's problem-solving process. Each student may request multiple
hints, provided that the solution given by the student is not empty. The
goal of the hint is to either supplement the student's solution in cases
when the student is on the right path, but does not know how to
continue, or to offer a new partial solution in cases when the student
is moving away from the correct solution. Students have the option to
replace their query with the one from the hint or to ignore the hint
altogether. The system therefore acts as a query formulator, correcting
students' queries when requested. Hints are formed using solutions from
the previous generations of students. The solutions were collected
during years \(2012\) and \(2014\) in a process of evaluating student's
SQL skills. Overall the solution pool contained over \(30.000\) entries
spread across approximately \(60\) exercises and \(2\) different
database schemas (including the well-known northwind database). Each
entry was described with the query in plain text format, the final score
of the query, user, schema, exercise id and time.

Figure \ref{fig:architecture} depicts all three major perspectives of
the system -- the teacher's view, the student's view and the system's
view. The process of SQL learning starts with an instructor preparing
the exercises to be used for evaluation (assignment formulation). The
evaluation rules and ideal solutions need to be provided along with a
description of the task and with a representative image of the database
schema. Evaluation rules determine the type of scoring used for each
exercise, as some exercises may require the student to return rows in a
specific order, to name columns in a specific way, etc. Ideal solutions
are necessary since exercises may not have matching past (correct)
solutions to rely on for hint generation. They also help address the
cold-start problem mentioned later on. In our evaluation, exercises were
formulated with unambiguity in mind, that is, there is a single unique
result matrix for every exercise, which is \(100\%\) correct. Several
ideal solutions were provided per exercise, each returning the same
ideal result matrix, but using a different concept of data retrieval
(e.g.~joining multiple tables, aggregating data and nesting queries). In
the second stage, the students solve SQL exercises. Their actions are
recorded and their solutions used as data for future generations of hint
consumers. During the course of exercise-solving process, the students
have the option to test the query and receive the result matrix or to
request a hint, which then augments their solution. Once satisfied, the
students submit their solution as final.

The basis of the system's perspective is a set of past attempts
(queries) at solving SQL exercises, provided by the students from
earlier generations. The historical data contain enough information to
create a model, which is capable of generating useful hints. The system
utilizes MDPs to model the learning process. MDP is defined as a tuple

\begin{equation}
\langle \mathcal{S}, \mathcal{A}, \mathcal{P}, \mathcal{R}, \gamma \rangle
\label{eq:tuple}
\end{equation}

where \(\mathcal{S}\) is a finite set of states, \(\mathcal{A}\) a
finite set of actions connecting the states, \(\mathcal{P}\) a matrix of
transition probabilities, \(\mathcal{R}\) a reward function and
\(\gamma\) a discount factor. The system's behaviour is stochastic, that
is, the probability matrix defines, for each action, the probability
that this action will actually lead to the desired state. The actions
therefore have multiple destinations, with each destination being
reachable with a given probability. The reward function specifies, for
each state, the reward the agent receives upon reaching this state. Note
that the reward can also act as a punishment (negative values).
Furthermore, a policy \(\pi\) defines a mapping of states to actions.
Each policy completely defines the behaviour of an agent in the system
as it specifies which action the agent should take next, given the
current state. We can define a value function as

\begin{equation}
V^{\pi}(s) = \sum_{s'} {P_{ss'}(\mathcal{R}_s + \gamma V^{\pi}(s'))}
\label{eq:value-function}
\end{equation}

The goal of MDPs is to find an optimal policy, which maximizes the value
function -- the reward the agent receives in the future -- over all
policies, or equivalently \(V^*(s) = \max\limits_{\pi}(V^{\pi}(s))\). We
use value iteration to iteratively compute better estimations of the
value function for each state until convergence

\begin{equation}
V^{*}_{i+1}(s) = \max\limits_{a \in \mathcal{A}}\Big(\mathcal{R}_s + \gamma \sum_{s'}\mathcal{P}^{a}_{ss'} V^{*}_{i}(s')\Big)
\label{eq:value-iteration}
\end{equation}

The value iteration computes, for each state, its new value, given the
outgoing actions of the state. The new value of the current state is its
baseline reward (\(\mathcal{R}_s\)) plus the maximum contribution among
all actions leading to neighbours. A contribution from following a
specific action is equal to the weighted sum of the probability of
reaching the action's intended target and the value of the target
(neighbour) from the previous iteration. The discount factor gives
priority to either immediate or long-term rewards. Observing equation
\eqref{eq:value-iteration}, we can notice that when \(\gamma\) is low, the
contributions from neighbouring states drop with increasing number of
iterations.

\begin{figure}

{\centering \includegraphics[width=1\linewidth]{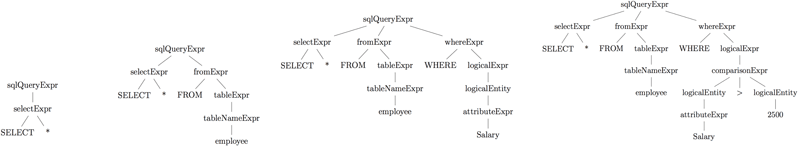}

}

\caption{Example of solution steps construction}\label{fig:solution-steps}
\end{figure}

Our historical data contain just the timestamp, user identification code
and submitted SQL query in plain text format. In order to close the gap
between raw data and high-level MDP states, we constructed our own
version of SQL language parser (component \texttt{A3} in Figure
\ref{fig:architecture}). The tool that proved useful was ANother Tool
for Language Recognition (ANTLR) \citep{parr_antlr_1995}. ANTLR takes as
an input the grammar, which defines a specific language (a subset of
context-free languages is supported by ANTLR), and generates parsing
code (Java code in our case). Through the use of grammar and lexer
rules, a parser was constructed that converts plain text SQL query into
a tree-like structure. An example of query parsing is evident from
Figure \ref{fig:solution-steps}, where internal tree nodes represent
expressions and leaves represent the actual (terminal) symbols of the
query.

\begin{figure}

{\centering \includegraphics[width=0.8\linewidth]{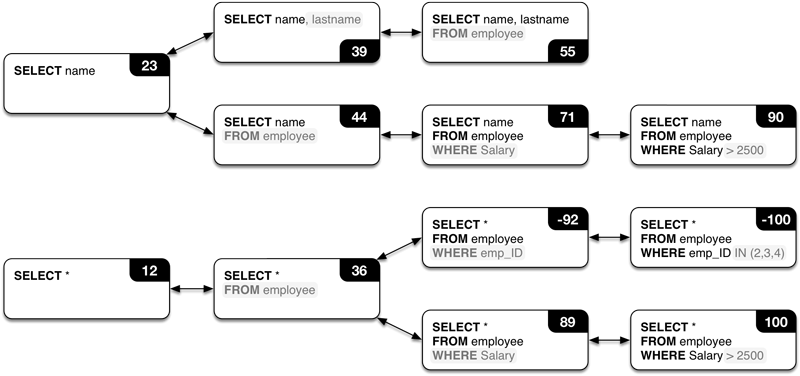}

}

\caption{Example of MDP graph. The numbers next to the states represent the rewards. The final states are in this case states with rewards $100$, $-100$, $90$ and $55$}\label{fig:MDP-graph}
\end{figure}

The individual solution steps, such as in Figure
\ref{fig:solution-steps}, are then directly mapped to states in MDP
notation. A single MDP state is therefore a tree. Actions are added
among consecutive solution steps while making sure there are no
duplicate states. We simplify the MDP construction by allowing each
state to have only one outgoing action, however that action leads to
multiple states. The probability matrix is calculated as the relative
frequency of users that have moved from a certain state to another with
regard to all users. The rewards are set only for the final states, that
is, the states that represent the final step of the solution. Reward
function is defined using the existing query evaluation component
(component \texttt{A4} in Figure \ref{fig:architecture}). When the query
score is high (\(>95\%\)), the state receives a high reward. In all
other cases, the state receives a ``negative reward''. This is to
prevent offering hints, which are only partially correct. The discount
factor was set to \(1,0\), giving priority to long-term rewards. In
addition, we add backward actions to all states, to aid the students
when they happen to be on the wrong path. Backward actions allow the
system to return from an incorrect solution path when the reward of
taking the forward action is worse than the reward of taking the
backward action. Furthermore, we seed the MDP with ideal solutions to
improve the hint generation process (component \texttt{A2} in Figure
\ref{fig:solution-steps}). This partially solves the cold-start problem
of new exercises, which do not yet have historical data available for
hints, as it allows the students to receive hints leading them to one of
the ideal solutions. The resulting MDP is a graph (not necessarily
connected). An example of such graph is visible in Figure
\ref{fig:MDP-graph}.

Let a branch denote one of the paths in one of the connected components
of MDP graph, which starts with the first solution step and ends with a
final solution. A single branch represents how a single student
constructed the solution. When a branch is split, that is, when at a
given node of the path/branch, that node leads to multiple other states,
then from this point forward, students deviated with their approach to
solving this exercise. In cases when the agent is located in an
incorrect sub-branch (a path leading to a final incorrect state), the
system returns from the incorrect subbranch to the first common ancestor
state of both the incorrect sub-branch and an alternative correct
branch. Skipping the entire incorrect sub-branch is necessary in order
to ensure that hints are progressive. For example, in Figure 3, if an
agent was located in a state corresponding to the use of IN clause
(state with reward \(-100\) in Figure \ref{fig:MDP-graph}), the system
would return two steps back and then offer a hint from there.

\begin{figure}

{\centering \includegraphics[width=0.8\linewidth]{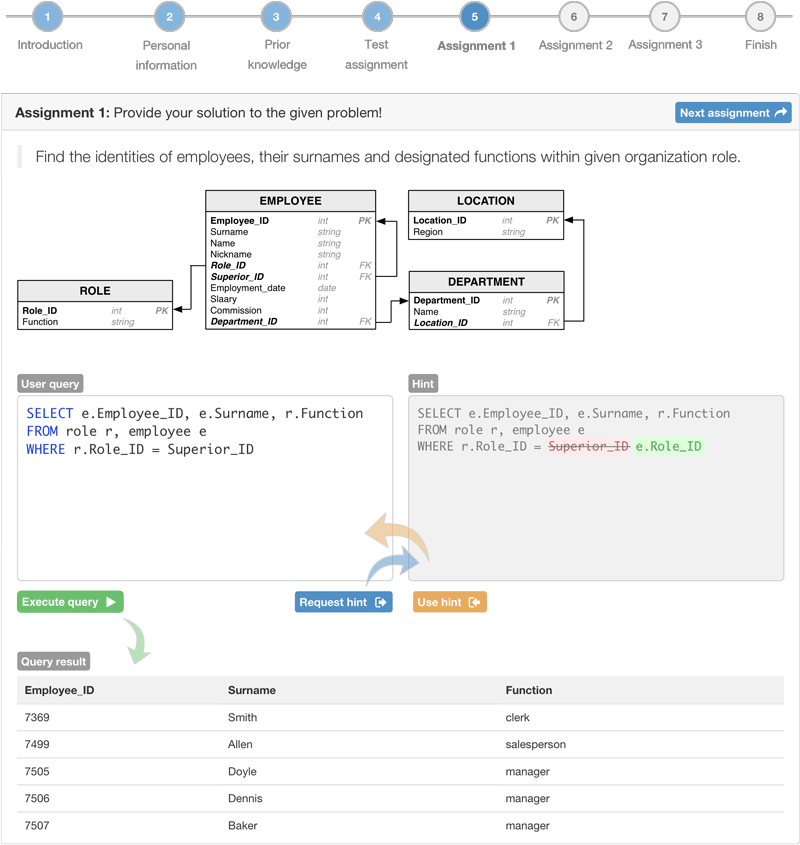}

}

\caption{Recommender system’s user interface}\label{fig:GUI}
\end{figure}

The hint generation process consists of the following activities (see
Figure \ref{fig:architecture}). When a student requests a hint, the
system accepts the current student's query as the input in order to
match it with one of the states in the MDP graph. The student's query is
parsed into a tree structure and then matched to the most similar state
(solution step) in the MDP (component \texttt{A5} in Figure
\ref{fig:architecture}). The MDP must be constructed first and because
this process is resource intensive (database I/O), we use an in-memory
cache (component \texttt{A7} in Figure \ref{fig:architecture}) to store
MDPs for each exercise. After the MDP is constructed, we apply value
iteration to determine the rewards of remaining, non-final states. Once
the MDP is retrieved and the matching state found, the hint is
constructed using the next best state given the matching state. This
includes converting the solution step into a text representation. An
example of hint construction is visible in Table
\ref{tab:hints-example}. The first row of Table \ref{tab:hints-example}
corresponds to a scenario, where the student is located in an incorrect
MDP branch. Observe that the system does not direct the student towards
the ideal solution (one of the ideal solutions actually), but proposes
an alternative MDP branch, which eventually leads to the correct
solution. The alternative MDP branch was constructed by another
student's solution from the previous generation. The last hint in the
first row demonstrates that nested queries are also supported. The
second row of Table \ref{tab:hints-example} corresponds to a scenario,
where the student's solution is partially correct, yet the student fails
to continue. As one can observe, the student forgot to include the
department table, which is what the hint corrects. A visual example of
how the hints are presented can be seen in Figure \ref{fig:GUI}.

\begin{table}

\caption{\label{tab:hints-example}Example of constructed hints for specific exercises, given student’s solution}
\centering
\begin{tabu} to \linewidth {>{\raggedright}X>{\raggedright}X>{\raggedright}X>{\raggedright}X}
\toprule
Task description & Ideal solution & Student's solution & Hints\\
\midrule
Return the number of employees in department "SALES" & \textbf{SELECT} COUNT(*) \textbf{FROM} employee, department \textbf{WHERE} employee.dept\_ID = department.dept\_ID AND department.name = "SALES" & \textbf{SELECT} * \textbf{FROM} department & \textbf{SELECT} COUNT(*) \textbf{FROM} department \textbf{WHERE} dept\_ID \textbf{SELECT} COUNT(*) \textbf{FROM} department \textbf{WHERE} dept\_ID IN (\textbf{SELECT} dept\_ID)\\
Return the number of employees in region "DALLAS" & \textbf{SELECT} COUNT(*) \textbf{FROM} employee e, department d, location l \textbf{WHERE} e.dept\_ID = d.dept\_ID AND d.loc\_ID = l.loc\_ID AND region = "DALLAS" \textbf{GROUP BY} region & \textbf{SELECT} COUNT(e.emp\_ID) \textbf{FROM} employee e, location l \textbf{WHERE} region = "DALLAS" & \textbf{SELECT} COUNT(e.emp\_ID) \textbf{FROM} employee e, location l, department d\\
\bottomrule
\end{tabu}
\end{table}

So far we have not mentioned how is the state matching actually
performed. Because all SQL queries are represented using tree
structures, we perform the matching using a tree distance criterion.
More specifically we employ the Zhang--Shasha algorithm
\citep{zhang_simple_1989}, which performs tree distance calculation for
ordered trees. There are several improvements we had to consider to make
the distance metric feasible. The SQL queries mostly contain aliases,
which help the user distinguish two instances of the same table. Because
aliases do not follow any syntax, they usually differ from user to user,
thus increasing the distance between trees. To alleviate the problem, we
perform alias renaming before matching the states. The process renames
all aliases to a common name, improving matching. In addition, the
Zhang--Shasha algorithm only performs distance calculation for ordered
trees. SQL queries are represented using unordered trees, as the order
of, for example, selected tables is irrelevant with respect to the final
solution. Tree distance for unordered trees is known to be an NP-hard
problem. Because trees representing queries are relatively small, we are
able to perform unordered tree distance calculation by treating children
of certain nodes as unordered sets and then defining a distance metric
for comparing the sets. The set distance calculation involves
recursively calculating the tree distance for each pair of elements from
both sets.

\subsection{Limitations}\label{limitations}

The strict matching of tree structures may in some scenarios cause
problems. Several small changes to the query can reduce the system's
ability to find a matching state and offer a hint. Even though we have
implemented several additional mechanisms to alleviate the problem
(alias renaming, unordered matching) there are still certain issues due
to inputs not following a specific syntax. A simple example is attribute
renaming (AS keyword), for exercises where the output needs to follow
specific naming rules. In general, we need to handle all cases of
free-form user input.

Another limitation of the system is the fact that all hints strictly
follow a predefined order of query construction. This is due to the lack
of solution steps and our assumption that students construct their query
in order of sections (\texttt{SELECT} clause first, then \texttt{FROM}
clause, etc.). The system, as a result, does not support hint delivery
for arbitrary order of query construction. Instead, the students are
expected to construct their queries by the sections of the query.

A potential problem is also the low reward of a certain state even
though the state might be a part of the correct solution path. The case
when a lot of students moved from a correct state to an incorrect path
(when there is an action with high probability leading to a state with
low reward) is reflected directly in the system's ability to offer a
hint. Such state will also have a low reward after value iteration even
though it is a part of a correct path. However, it is also a part of the
incorrect path and is more likely to lead to incorrect solutions. The
hints will, because of this, lead to other paths, leaving this state
unexplored.

All of the mentioned limitations do not have a major impact on the
learning outcome of the system and its ability to deliver hints. As we
show in the next section, the effectiveness of the system is more than
satisfactory.

\section{Evaluation}\label{evaluation}

\subsection{Method}\label{method}

The recommender system was implemented as a web application that
participants accessed using one of the modern web browsers. The process
of providing hints during solving the SQL assignments was evaluated on a
group of \(93\) participants, where each of them completed \(3\)
assignments, resulting in \(279\) solutions.

During the experiment, before participants started working on
assignments, they provided some information about their prior knowledge
(SQL proficiency level, years of experience in SQL) to enable clustering
and detailed analysis of participants. To obtain diversity of
participants' prior knowledge we employed \(60\) undergraduate students
at University of Ljubljana, Faculty of Computer and Information Science
with right-skewed prior knowledge and \(33\) more experienced
participants in using SQL with left-skewed prior knowledge, resulting in
a nearly normal distribution of participants' prior knowledge
(self-reported SQL proficiency and years of experience).

All of the participants' actions on the web site were recorded for
further analysis of the system employment. That allowed us to analyse
participant's time spent on reading instructions, solving the
assignment, time spent out of focus with the main window of the
assignment and the number of lost focuses. Furthermore, partial query
results, distance to correct solution, number of branches and if
participant is in the MDP branch, corresponding to a correct solution
path were also recorded to construct the participant timeline in solving
individual assignment.

Each participant was required to provide a solution to three randomly
selected SQL assignments. Each of the assignments was classified into a
category based on the difficulty level, where every participant was
randomly allocated one easy (selection of attributes from one table and
filtering with simple predicates), one moderate (using join to merge
data from multiple tables and filtering with more advanced predicates)
and one difficult (using nested queries, grouping and aggregation
functions) assignment.

As Figure \ref{fig:GUI} depicts, every assignment consisted of
instructions, a graphical view of the conceptual model, a query box for
entering the solution, an optional hint box and interactive results of
the user's query.

Participants generally first enter a query into the user's query box and
interactively evaluate the results by pressing the ``Execute query''"
button. At some point, if the participants were unable to continue, they
requested a hint from the recommender system, which was then displayed
next to the user query with indicated adaptations of the current query
(added elements in green colour and removed elements in red colour) (see
Figure \ref{fig:GUI}). If participants found the hint beneficial, they
could instantly employ it by clicking the ``Use hint'' button. Then the
participant could again check the results of their current query by
executing it. When one was satisfied with the result, they could
continue to the next assignment. If hints were employed, then the
participant indicated the level of suitability of provided hints in
additional questionnaire.

Additionally, the group of \(60\) undergraduate students were presented
with a direct encouragement - their final result (average score of three
random assignments) was considered as a part of study obligations of
their degree programme, while \(33\) more experienced participants were
only asked to participate for the evaluation purposes.

Every participant could request unlimited number of hints per
assignments. To discourage participants to excessive use of hints or
even solving the complete assignment with hints only, a small score
penalty (inversely proportional to assignment complexity) was introduced
for hint employment, which was clearly introduced to the participants
before starting the evaluation.

After outlier detecion based on recorded time (instructions reading,
solving and unfocussed) and observation of participant's timelines,
\(37\) attempts were excluded from the initial data set. Further
analysis of our results was conducted on \(242\) attempts, where we
observed the following assignment solving dynamics (including \(95\%\)
bootstrapped confidence intervals). The mean time participants spent on
reading the assignment's instructions was \(9s\) \([7s, 10s]\), while
the mean time spent on solving the assignment was \(268s\)
\([242s, 294s]\). The mean number of requested hints per assignment was
\(1,43\) \([1,08, 1,82]\), while \(82\) out of \(242\) (\(33,88\%\))
assignments was solved using at least one hint.

\subsection{Results and discussion}\label{results-and-discussion}

To evaluate the performance of proposed recommender system, we cluster
participants into five segments, based on prior knowledge and feedback
on hint usefulness:

\begin{itemize}
\tightlist
\item
  \textbf{Segment I}: attempts of knowledgeable participants without
  employing hints,
\item
  \textbf{Segment II}: attempts of not knowledgeable participants
  without employing hints,
\item
  \textbf{Segment III}: attempts of not knowledgeable participants, not
  finding hints useful,
\item
  \textbf{Segment IV}: attempts of knowledgeable participants, finding
  hints useful and
\item
  \textbf{Segment V}: attempts of not knowledgeable participants,
  finding hints useful.
\end{itemize}

When observing an individual participant in solving a given assignment,
several actions are recorded (e.g.~altering current solution, executing
query, requesting a hint, employing a hint, losing a focus of the
current window with the assignment, etc.). With evaluation of the
proposed recommender system, we focus mainly on \emph{\textbf{the
distance from the correct solution}} \(\boldsymbol{dist_{sol}}\) in a
given time. The distsol is defined as the minimum number of solution
steps required reaching the correct state in the MDP graph, given a
current state. The metric is calculated using a simple distance
criterion, where states are treated as nodes in a graph and actions as
links between states. The breadth-first traversal is used to determine
the shortest distances and the minimum of all distances is taken.

We define a linear association

\begin{equation}
\widehat{dist_{sol}} = \alpha +\beta_{type} \cdot \widehat{t}_{elapsed}
\label{eq:dist-sol}
\end{equation}

where
\(\widehat{dist_{sol}} = \frac{dist_{sol}}{\left \| dist_{sol} \right \|}\),
\(\widehat{t}_{elapsed} = \frac{t_{elapsed}}{\left \| t{elapsed} \right \|}\)
and \emph{type} \(\in\) \emph{(all, pre\_first\_hint, post\_first\_hint,
after\_hint\_avg)}.

When analysing the results, we focus on the following performance
indicators:

\begin{itemize}
\tightlist
\item
  participant's solving time per assignment,
\item
  number of different branches participant encounters per assignment and
\item
  various regression coefficients of participant timeline, when solving
  the given assignment:

  \begin{itemize}
  \tightlist
  \item
    \(\beta_{all}\) is based on all participant's actions,
  \item
    \(\beta_{pre\_first\_hint}\) or \(\beta_{pre\_fh}\) is based on
    actions before the first hint employment,
  \item
    \(\beta_{post\_first\_hint}\) or \(\beta_{post\_fh}\) is based on
    actions after the first hint employment and
  \item
    \(\beta_{after\_hint\_avg}\) or \(\beta_{aha}\) is based on
    consequent actions after each hint employment, with an average over
    all hint requests.
  \end{itemize}
\end{itemize}

The following Table \ref{tab:participant-segmentation} depicts
aggregated mean results per predefined five participants' segments. When
examining the results, we endeavour to obtain as negative b values as
possible, which indicate rapid advancement towards correct solution
(e.g.~minimize distance to correct solution over time) and by doing that
measure the effect of hint employment.

\begin{table}

\caption{\label{tab:participant-segmentation}Participant segmentation}
\centering
\begin{tabu} to \linewidth {>{\centering}X>{\raggedright}X>{\centering}X>{\raggedleft}X>{\raggedleft}X>{\raggedleft}X>{\raggedleft}X>{\raggedleft}X}
\toprule
Segment & Prior knowledge & Hints useful & $\boldsymbol{t_{solving}}$ & $\boldsymbol{n_{branches}}$ & $\boldsymbol{\beta_{pre\_fh}}$ & $\boldsymbol{\beta_{aha}}$ & $\boldsymbol{\Delta\beta}$\\
\midrule
I & Knowledgeable & -- & $228,00$ & $3,88$ & -- & -- & --\\
II & Not knowledgeable & -- & $254,17$ & $4,44$ & -- & -- & --\\
III & Not knowledgeable & No & $381,76$ & $4,65$ & $-0,47$ & $-5,67$ & $-5,20$\\
IV & Knowledgeable & Yes & $243,00$ & $3,60$ & $-2,00$ & $-6,36$ & $-4,27$\\
V & Not knowledgeable & Yes & $322,80$ & $4,42$ & $-0,85$ & $-10,76$ & $-9,89$\\
\bottomrule
\end{tabu}
\end{table}

When observing regression coefficients of participant timeline or the
degree of convergence to the final solution, we focus on
\(\beta_{pre\_first\_hint}\), \(\beta_{after\_hint\_avg}\) and
\(\Delta\beta = \beta_{after\_hint\_avg} - \beta_{pre\_first\_hint}\) to
measure the impact of hint employment. The aforementioned metrics can be
calculated only for segments III, IV and V, as participants from
segments I and II did not employ hints (see Table
\ref{tab:participant-segmentation}).

\begin{figure}

{\centering \includegraphics[width=0.5\linewidth]{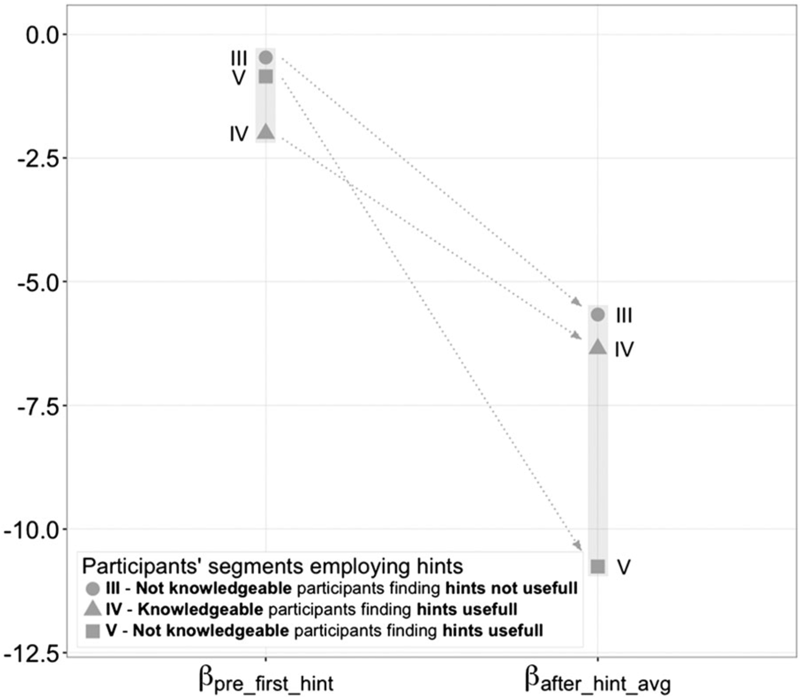}

}

\caption{Degree of convergence to the final solution $\beta$}\label{fig:degree-of-convergence}
\end{figure}

Figure \ref{fig:degree-of-convergence} depicts a sharp decline in the
degree of convergence to the final solution \(\beta\) before and after
the hint employment for all segments of participants that employed hints
(III, IV and V). The finding is also statistically confirmed by
Mann--Whitney--Wilcoxon test for individual segments:

\begin{itemize}
\tightlist
\item
  \(\beta^{III}_{after\_hint\_avg} = -5,67 < \beta^{III}_{pre\_first\_hint} = -0,47\)
  with \(W^{III} = 73\) and \(p^{III} = 7,2 \times 10^{-3}\),
\item
  \(\beta^{IV}_{after\_hint\_avg} = -6,36 < \beta^{IV}_{pre\_first\_hint} = -2,00\)
  with \(W^{IV} = 60\) and \(p^{IV} = 2,6 \times 10^{-2}\) and
\item
  \(\beta^{V}_{after\_hint\_avg} = -10,75 < \beta^{V}_{pre\_first\_hint} = -0,85\)
  with \(W^{V} = 468\) and \(p^{V} = 9,8 \times 10^{-8}\).
\end{itemize}

We can conclude that the hint, provided by the recommender system, is
statistically significant in terms of the impact on the degree of
convergence to the final solution as \(\beta_{after\_hint\_avg}\) values
of all three segments are significantly lower than
\(\beta_{pre\_first\_hint}\).

\begin{figure}

{\centering \includegraphics[width=0.5\linewidth]{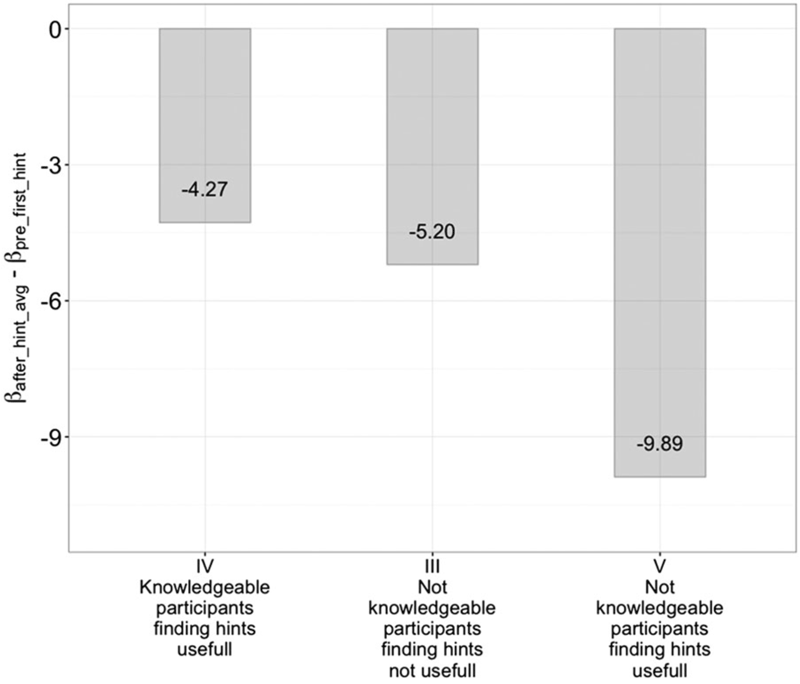}

}

\caption{Impact of hint employment}\label{fig:hint-impact}
\end{figure}

When considering the magnitude of hint employment impact from Figure
\ref{fig:degree-of-convergence}, we can observe
\(\Delta\beta = \beta_{after\_hint\_avg} - \beta_{pre\_first\_hint}\) in
Figure \ref{fig:hint-impact}. The results show that participants in
segments V, III and IV respectively found the provided hints most
beneficial. The biggest impact of hint employment on the degree of
convergence to the final solution Db can be observed in not
knowledgeable participants, finding hints useful (segment IV), which is
what was also expected prior evaluation, under assumption that the
recommender system would be effective. Similar effects can be observed
with knowledgeable participants, finding hints useful (segment V), but
the effect size is lower, but still significant. More intriguing finding
is that the positive impact of hints employment can also be observed in
not knowledgeable participants, not finding hints useful (segment III).
Even though participants indicated that hints are not useful, the
results show that their advancement towards correct solution, after
employing hints was significantly better than before employing hints.

\begin{figure}

{\centering \includegraphics[width=0.5\linewidth]{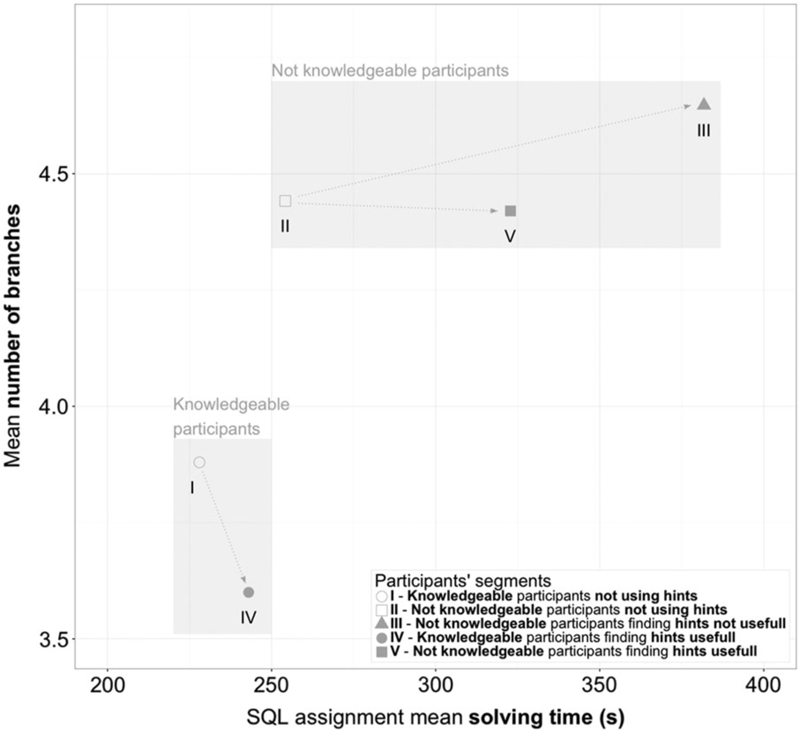}

}

\caption{Impact of prior knowledge in hint employment}\label{fig:prior-knowledge-in-hint-employment}
\end{figure}

To examine the impact of prior knowledge, even further we studied the
impact of prior knowledge in hint employment considering assignment
solving time (\(t_{solving}\)) and number of different branches
(\(n_{branches}\)) in Figure
\ref{fig:prior-knowledge-in-hint-employment}. Assignment solving time
includes only time designated to solving (without instructions reading
time and unfocussed time), while number of different branches defines
the number of branches participant was designated for in MDP, when
solving the SQL assignment. The higher values of nbranches indicate
participant exploring significantly different paths to the correct
solution and usually not knowing how to continue.

Figure \ref{fig:prior-knowledge-in-hint-employment} depicts that we can,
based on \(t_{solving}\) and \(n_{branches}\), cluster together segments
I and IV (knowledgeable participants) and segments II, III and V (not
knowledgeable participants). The general observation is that
knowledgeable participants on average tend to spend less time solving
SQL assignments, and are related to lower number of branches than not
knowledgeable participants. The arrows in Figure
\ref{fig:prior-knowledge-in-hint-employment} indicate the influence of
hint employment. We can observe that when knowledgeable participants
(segments I an IV) employ hints, \(t_{solving}\) increases but the
\(n_{branches}\) decreases, indicating that the number of different
paths to the correct solution drops, resulting in more consistent
progressing towards final solution. In case of not knowledgeable
participants (segments II, III and V), the impact of hints increases
\(t_{solving}\), while \(n_{branches}\) slightly decreases when
participants find hints useful (segment V) and increases when
participants do not find hints useful (segment III).

\begin{figure}

{\centering \includegraphics[width=1\linewidth]{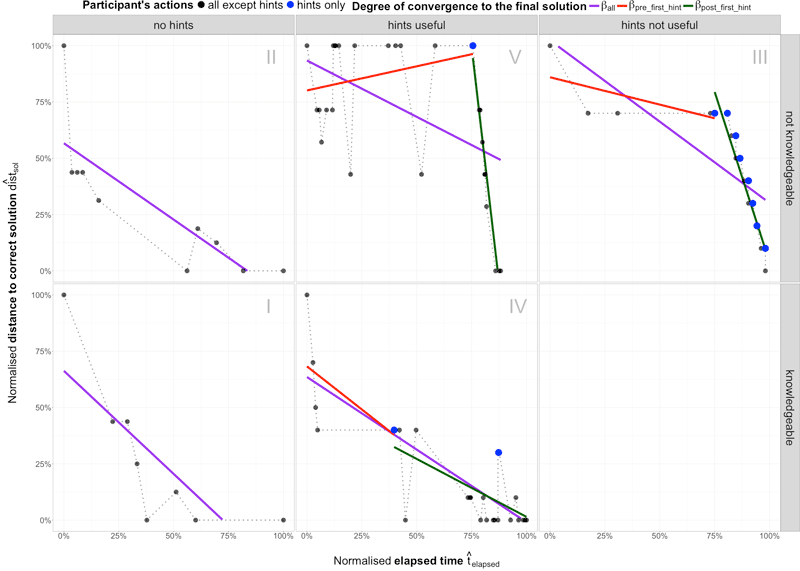}

}

\caption{Medoids of participant’s timeline}\label{fig:timeline-medoids}
\end{figure}

To further investigate the differences between five segments of
participants, Figure \ref{fig:timeline-medoids} details the most
representational participant of every cluster segment that is the most
similar to the median values of the characteristics of performance
indications \(t_{solving}\), \(n_{branches}\),
\(\beta_{pre\_first\_hint}\), \(\beta_{after\_hint\_avg}\) and
\(\Delta\beta\). Figure \ref{fig:timeline-medoids} depicts the timeline
of representational participant solving the assignment with every
recorded action, where hint employment is highlighted (blue dots).
Additionally, three regression lines and coefficients are also depicted,
based on the following filtering:

\begin{itemize}
\tightlist
\item
  \(\beta_{all}\) for all actions,
\item
  \(\beta_{pre\_first\_hint}\) for actions before first hint employment
  and
\item
  \(\beta_{post\_first\_hint}\) for actions after the first hint
  employment.
\end{itemize}

When considering segment I (knowledgeable participants without employing
hints) and segment II (not knowledgeable participants without employing
hints), we can observe that both groups did not employ hints, hence only
ball is depicted. The average participant in segment II is not
knowledgeable and he experienced more fluctuations in his path to the
final solution than knowledgeable participant in segment I.

When observing participants in segment III (not knowledgeable
participants, not finding hints useful), we can conclude that the
average participant is able to solve the SQL assignment to certain
degree, but then does not know how to continue. At that point, the
participant starts requesting hints, which he/she does not find useful.
Nevertheless, the hints still lead participant's current solution to the
correct final solution that is the most similar to his/her previous
steps in the attempt.

We would like to emphasize attempts in segment IV (knowledgeable
participants, finding hints useful) and segment V (not knowledgeable
participants, finding hints useful) that both include attempts where
participants found hints useful and recommender system provides the most
beneficial results, as the impact of hint employment is evident. The
average knowledgeable participant from segment IV is able to solve the
SQL assignment but is at some points uncertain whether he/she is on the
correct path. In that case he/she employs hints, which confirm his/her
previous steps and guide him/her to the correct final solution. The
average not knowledgeable participant from segment V is at some point
unable to continue and after hint employment he/she experience the boost
in terms of rapid decline of a distance to correct solution
\(\widehat{dist_{sol}}\).

\section{Conclusions and future work}\label{conclusions-and-future-work}

In this study, we set out to construct a system to help students learn
SQL. The system makes use of past student attempts at solving
SQL-related exercises. We employed MDPs to encode the knowledge from
historical data and to traverse the states to find the most suitable
hint. We parse the SQL language and generate the solution steps in order
to close the gap between raw queries and MDP states. We also seed the
system with additional expert solutions to improve our ability to
deliver a hint.

We tested our system in an actual learning environment. The results
indicate that the hints are well accepted and drastically reduce the
distance to correct solution even after several steps upon using the
hint. If a student's distance to correct solution alternates before
requesting a hint, it is more stable after receiving a hint. The goal of
the hints is not merely to improve the overall score of the student, but
has a broader intent. If a student does not know how to proceed, they
receive a hint, which in turn proposes a new approach to solving a
problem. The students therefore explore alternative paths, which they
would not consider on their own. This is also why initially, after
receiving a hint, some students still alternate with regard to the
distance to correct solution, as they are exploring the unfamiliar
states and perform errors doing so. As expected, the system has the
largest impact on students with low prior knowledge, which is desired.

Our system would benefit from certain improvements in the future. An
obvious improvement might be to perform matching of the states not
through comparison of tree structures, but rather using a more
high-level approach. We could, for example, detect key concepts from
every query and then count the number of common concepts whilst
comparing two queries. This would also allow discarding any free-form
user input from comparison. Another improvement we will consider in the
future is to abandon the assumption that the students solve the problems
by the order of sections of the query. In order to do that, we would
first have to gather new historical data, which also contains steps of
query construction. Afterwards, hints could be employed for an arbitrary
query construction permutation.

\end{document}